\documentclass[10pt,letterpaper]{article}
\usepackage[top=0.85in,left=1.65in,footskip=0.75in]{geometry}

\usepackage{amsmath,amssymb}

\usepackage{changepage}

\usepackage[utf8x]{inputenc}

\usepackage{textcomp,marvosym}

\usepackage{cite}

\usepackage{nameref,hyperref}

\usepackage[right]{lineno}

\usepackage{microtype}
\DisableLigatures[f]{encoding = *, family = * }

\usepackage[table]{xcolor}

\usepackage{array}

\usepackage{siunitx}

\usepackage{csquotes}

\usepackage{nicefrac}

\newcolumntype{+}{!{\vrule width 2pt}}

\newlength\savedwidth

\usepackage[aboveskip=1pt,labelfont=bf,labelsep=period,justification=raggedright,singlelinecheck=off]{caption}


\date{}

\usepackage{lastpage,fancyhdr,graphicx}
\usepackage{epstopdf}
\newcommand{\Fig}[1]{Fig~\ref{fig:#1}}

\newcommand{\R}{\mbox{\rm \hbox{I\kern-.15em\hbox{R}}}}

\renewcommand{\vec}[1]{\mathbf{#1}}
\newcommand{\mat}[1]{\mathbf{#1}}
\newcommand{\laplace}{\Delta}
\newcommand{\set}[1]{\mathcal{#1}}
\newcommand{\of}[1]{\!\left( #1 \right)}
\newcommand{\abs}[1]{\left| #1 \right|}
\newcommand{\norm}[1]{\left\Vert {#1} \right\Vert}

\newcommand{\Efit}{E_{\mathrm{fit}}}

\newcommand{\Ereg}{E_{\mathrm{reg}}}
\newcommand{\Epca}{E_{\mathrm{PCA}}}

\newcommand{\lreg}{\lambda_{\mathrm{reg}}}
\newcommand{\ltik}{\lambda_{\mathrm{tik}}}
\newcommand{\mean}[1]{\mathbf{\bar{#1}}}

\newcommand{\noSkullCTs}{60}
\newcommand{\noSkullScans}{2}
\newcommand{\noSkulls}{62} % adjust if number of CTs or scans changes
\newcommand{\noHeadCTs}{43} % proofed! all CTs for the statistical evaluation can be used for the face model!
\newcommand{\noHeadScans}{39}
\newcommand{\noHeads}{82} % adjust if number of CTs or scans changes
\newcommand{\noCTs}{43}

\begin{document}
\thispagestyle{empty}
%\vspace*{0.2in}

\begin{flushleft}
{\Large
     \textbf\newline{A method for automatic forensic facial reconstruction based on dense statistics of soft tissue thickness}
    % Please use "sentence case" for title and headings (capitalize only the
    % first word in a title (or heading), the first word in a subtitle (or
    % subheading), and any proper nouns).
}
\newline
\\
Thomas Gietzen\textsuperscript{1},
Robert Brylka\textsuperscript{1},
Jascha Achenbach\textsuperscript{2},
Katja zum Hebel\textsuperscript{3},
Elmar Schömer\textsuperscript{4},
Mario Botsch\textsuperscript{2},
Ulrich Schwanecke\textsuperscript{1},
Ralf Schulze\textsuperscript{5}
\\
\bigskip
\textbf{1} RheinMain University of Applied Sciences, Computing for New Media,
Wiesbaden, Germany
\\
\textbf{2} Bielefeld University, Computer Graphics Group, Bielefeld, Germany
\\
\textbf{3} University Medical Center of the Johannes Gutenberg University
Mainz, Department of Prosthetic Dentistry, Mainz, Germany
\\
\textbf{4} Johannes Gutenberg University Mainz, Institute of Computer Science,
Mainz, Germany
\\
\textbf{5} University Medical Center of the Johannes Gutenberg University Mainz, Section of Oral Radiology, Mainz, Germany
\bigskip

\end{flushleft}

%\tableofcontents
%\newpage
%\listoffigures
%\newpage

%%%%%%%%%%%%%%%%%%%%%%%%%%%%%%%%%%%%%%%%%%%%%%%%%%%%%%%%%%%%%%%%%%%%%%%%%%%%%%

\section*{Abstract}

In this paper, we present a method for automated estimation of a human face
given a skull remain. The proposed method is based on three statistical models.
A volumetric (tetrahedral) skull model encoding the variations of different
skulls, a surface head model encoding the head variations, and a dense
statistic of facial soft tissue thickness (FSTT). All data are automatically
derived from computed tomography (CT) head scans and optical face scans. In
order to obtain a proper dense FSTT statistic, we register a skull model to
each skull extracted from a CT scan and determine the FSTT value for each
vertex of the skull model towards the associated extracted skin surface. The
FSTT values at predefined landmarks from our statistic are well in agreement
with data from the literature.

To recover a face from a skull remain, we first fit our skull model to the
given skull. Next, we generate spheres with radius of the respective FSTT value
obtained from our statistic at each vertex of the registered skull. Finally, we
fit a head model to the union of all spheres. The proposed automated method
enables a probabilistic face-estimation that facilitates forensic recovery even
from incomplete skull remains. The FSTT statistic allows the generation of
plausible head variants, which can be adjusted intuitively using principal
component analysis. We validate our face recovery process using an anonymized
head CT scan. The estimation generated from the given skull visually compares
well with the skin surface extracted from the CT scan itself.

\linenumbers

%%%%%%%%%%%%%%%%%%%%%%%%%%%%%%%%%%%%%%%%%%%%%%%%%%%%%%%%%%%%%%%%%%%%%%%%%%%%%%
\nolinenumbers

\section*{Introduction}

Facial reconstruction is mainly used in two principal branches of science:
forensic science and anthropology. Remains of a human skull act as input to
reconstruct the most likely corresponding facial appearance of the dead person
to enable recognition. Traditional methods rely on manual sculpturing a
moldable substance onto the replica of the unknown skull using anatomic clues
and reference data. In her comprehensive review, Wilkinson~\cite{Wilkinson2010} considers this a
highly subjective procedure requiring a great deal of artistic interpretation
and thus providing rather unreliable results. For forensic investigations, a
reliable most-likely face estimate is demanded \cite{Wilkinson2010}.
Computer-based methods can provide consistent and objective results and also
allow the integration of meta-information, such as age, sex, or weight
\cite{Claes2006}.

Computer-aided facial reconstruction methods have been previously proposed in
other publications~\cite{Turner2005,Tu2007,Romeiro2014,Shui2016,Shui2017}.
Related work uses different techniques for the underlying registration as well
as for the subsequent facial reconstruction. Although not standardized, FSTT measurements play an important role both in facial approximation
and craniofacial superimposition methods due to the quantitative information
provided~\cite{Stephan2008}. A wide variety of different techniques such as
needle probing, caliper or radiographic measurements, or ultrasonographic
assessments are used to determine the FSTT, which lead to different results in
the FSTT statistics.  In addition, 3D imaging techniques
such as CT or Magnetic Resonance Imaging (MRI) are
employed for this purpose.  Driven by the generally lower radiation dose when
compared to medical CT, lately Cone Beam Computed Tomography (CBCT) has also
been used~\cite{Hwang2015}.
In general it is difficult to compare FSTT studies based on CT and CBCT scans.
CT scans are taken in supine position whereby CBCT scans can be taken in
various positions (sitting, lying down, standing up), which has different gravity effects on the FSTT.
CBCT also has the inherent drawback that some landmarks cannot be found in the
data sets because it is normally limited to the craniofacial region.
Although not backed by numerical data, it is
generally advocated to prefer measurements on living individuals over
cadavers~\cite{Stephan2008}. In \cite{Stephan2008}, Stephan and Simpson conclude that regardless of the
applied technique the measurement error for FSTT assessment is rather high
(relative error of around 10\%) and that no method so far can be considered
superior to any other. In addition, the authors stated that small sample sizes for most of the
studies also compromise the degree to which the results from such studies can
be generalized. 

Generally spoken, measurements based on a few distinct landmark
points yield the inherent drawback of providing only a few discrete thickness
values. Areas between these distinct measurement points need to be
interpolated. A \emph{dense} soft tissue map would yield important information
for facial reconstruction. A statistical head model could be fitted to such a
dense soft tissue profile thereby providing an estimate of the
visual appearance of the person to be identified, based on
\emph{statistics} of the sample data.

Turner et al.~\cite{Turner2005} introduced a method for automated skull
registration, and craniofacial reconstruction based on extracted surfaces from
CT data that was applied to a large CT data base consisting of 280 individuals
in \cite{Tu2007}. For registration of a known skull to a questioned one, the
authors use a heuristic to find crest lines in combination with a two-step ICP
registration followed by a thin-plate spline warping process. The same warping
function is applied to the extracted skin of the known skull.  Following, from
a collection of 50 to 150 warped skin surfaces they use principal component
analysis (PCA) to construct a \enquote{face-space} with a mean face for the
questioned skull. Using the linear combination of the eigenvectors with some
a-priori knowledge, such as age and sex, they are able to generate a subset of
most likely appropriate appearances for the questioned subject. To this end,
both the questioned and the known skull are represented as polygonal meshes and
are reduced to their single, outer surface. Thereby, disregarding the
volumetric nature of the bony structure in some cases leads to poor fitting
results.

The utilization of a deformable template mesh for forensic facial
reconstruction was presented by Romeiro et al.~\cite{Romeiro2014}.  Their
computerized method depends on manually identifying 57 landmarks placed on the
skull. Based on these preselected landmarks and a corresponding FSTT (obtained
from other studies) an implicit surface is generated using Hermite radial basis
functions (HRBF). To improve the quality of the result, they use several
anatomical rules such as the location of the anatomical planes and anatomical
regressions related to the shape of the ears, nose, or mouth.  Hence, the
quality of their results strongly depends on an appropriate template that
properly takes age, sex, and ethnicity into account.

An approach for craniofacial reconstruction based on dense FSTT statistics,
utilizing CT data, was presented by Shui et al.~\cite{Shui2016}.  Their method
depends on 78 manually selected landmarks placed on the skull, which guide the
coarse registration of a template skull to each individual skull, followed by a
fine registration using ICP and thin plate splines (TPS). The FSTT measurement
is performed for each vertex of the deformed skull in the direction defined by
the geometric coordinate. A coarse reconstruction of a face from an
unidentified skull is achieved by translating each skull vertex in the defined
direction by the length of the FSTT measured at this position.  To achieve a
smooth appearance six additional points have to be marked manually for guiding
a TPS deformation of a template face to the coarse reconstruction.  Finally,
the recovery of mouth, eyes, and nose has to be performed by a forensic
expert, which makes the method not fully automatic.

Shui et al.~\cite{Shui2017} proposed a method for determining the craniofacial
relationship and sexual dimorphism of facial shapes derived from CT scans.
Their approach employs the registration method presented in~\cite{Shui2016}, to
register a reference skull and face to a target skull respective face. Applying
a PCA to the sets of registered skull and skin templates, they derive a
parametric skull and skin model. Through analyzing the skull- and skin-based
principal component scores, they establish the craniofacial relationship
between the scores and therefore reconstruct the face of an unidentified
subject. Although the visual comparison of the estimated face with the real
shows good results, these results appear to be due to over-fitting. Moreover,
the geometric deviation, especially in the frontal part of the
face, are mostly around 2.5--5\,mm, which indicates rather inaccurate
reconstruction results.

Our approach to forensic facial reconstruction is divided into two parts: model
generation and forensic facial reconstruction. Unlike most previous
methods\cite{Turner2005,Tu2007,Romeiro2014,Shui2016,Shui2017} our approach is
fully automated, from the initial skull registration up to the final face
reconstruction, and thus does not require any manual interaction. Only the
initial model generation (preprocessing or training phase) requires a few
manual steps. The next section describes the generation of the three models
required for our automated facial reconstruction approach: The parametric skull
model, the statistic of FSTT, and the parametric head model. In the following
sections the automated facial reconstruction process is presented, including the
modeling of variants of plausible FSTT distributions for a given skull.

%%%%%%%%%%%%%%%%%%%%%%%%%%%%%%%%%%%%%%%%%%%%%%%%%%%%%%%%%%%%%%%%%%%%%%%%%%%%%%

\section*{Model generation}

\begin{figure}[htbp]
\centering
\includegraphics[width=\textwidth]{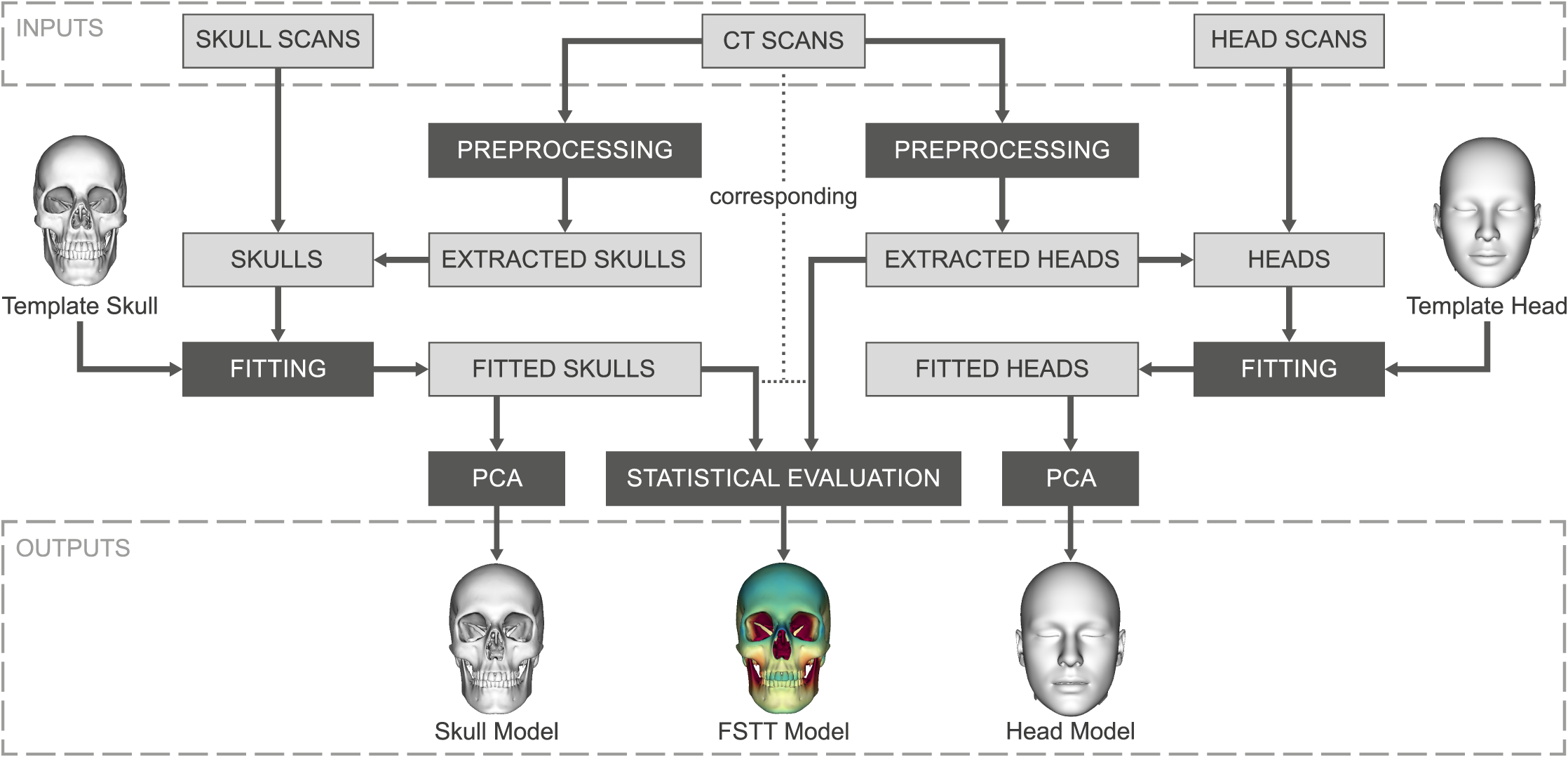}
\caption{\textbf{Overview of our model generation processes.} Generation of
a skull and a head model as well as a dense FSTT statistic from multimodal
input data.}
\label{fig:Fig1}
\end{figure}

In this section we present the proposed model generation processes, as outlined
in \Fig{Fig1}. We use volumetric CT scans and optical 3D surface scans as input
and distinguish between two input types: skulls and heads. 
In the following, the outer skin surface of a head is referred to as \emph{head} and the bony skull structure is referred to as \emph{skull}.
In order to obtain a uniform data basis, a \emph{preprocessing} step is performed to extract the skull and the head as triangular surface meshes from each CT scan.
%It is represented by a triangular surface mesh. The bony skull structure is referred to as \emph{skull} and represented by a tetrahedral volume mesh.  
In the next step we need to establish the relationship between different skulls as well as between different heads.
For this purpose, in a \emph{fitting process}, we register an appropriate template model to each given mesh of a specific input type. 
After that, we are able to utilize the fitted templates to determine the geometric variability of the skulls respectively heads performing a \emph{PCA}.
As result we derive two parametric models:
a parametric skull model and a parametric head model. 
Based on
corresponding skulls and heads extracted from CT scans we additionally build a
dense FSTT map in the \emph{statistical evaluation} step. 

%-----------------------------------------------------------------------------

\subsection*{Database}

Following internal ethical review board approval\footnote{Ethik-Kommission der
Landesärztekammer Rheinland-Pfalz, No 837.244.15 (10012)}, head CT scans were
collected from the PACS system of the University Medical Center Mainz.
We only used existing CT data (from four different CT devices) from our database. No subject was exposed to ionizing radiation for this research.
The local ethical approval board (Landesärztekammer Rheinland-Pfalz, Deutschhausplatz 2, 55116 Mainz) has approved the processing of the pseudonymized existing CTs (from the DICOM database of the University Medical Center Mainz) to generate the statistical models under the approval number No 837.244.15 (10012) (date: 05.08.2015).
In our study we included CT scans that meet the following criteria:
\begin{enumerate}
\item The facial skull of the patient is \emph{completely
imaged}.
\item The \emph{slice thickness} is less than or equal to \SI{1}{mm}.
\item The subject has no significant oral and maxillofacial deformations or missing parts.
\end{enumerate}

From several hundred CT scans that we analyzed a total number of \noSkullCTs\
were suitable for our purpose.
However, only \noCTs\ of these scans could be
used for generating the parametric head model and the statistic of FSTT, since
in the remaining 17 CT scans external forces (e.g.\ frontal extending neck
stabilizers, nasogastric tubes, etc.) compressed the soft tissue. In a
\emph{preprocessing} step every CT scan was cropped, such that we obtain a
consistent volume of interest limited to the head area. For this purpose the
most posterior point of the mandibular bone was determined automatically in the
2D slice images and the volume was trimmed with an offset below this detected
position. After this cropping step, bone and skin surface meshes were extracted
using the Marching Cubes algorithm~\cite{Lorensen1987} (we used the Hounsfield
units -200 and 600 as iso"=values for skin and bone surface extraction,
respectively). To remove unwanted
parts, such as the spine or internal bone structures, a connectivity filter was
applied to the bone mesh, leaving only the skull. Finally, all extracted meshes
were decimated to obtain a uniform point density for all data sets~\cite{Garland1997}. The meshes
extracted from CT data were supplemented by triangle meshes from 3D surface
head scans\footnote{From \url{www.3dscanstore.com}} of real subjects in order
to fill up the database for our model generation processes. The 3D surface
scans are of high quality, do not suffer from artifacts or strong noise, and
consist of about $500\,\mathrm{k}$ vertices in case of the head and about
$400\,\mathrm{k}$ vertices in case of the skull.  In summary the following
data sets were included in the study:

\begin{enumerate}

    \item A total number of $p=\noSkulls$ skulls
        (\noSkullCTs~extracted skulls from CT scans and \noSkullScans~skulls
        from 3D surface scans) were used to generate a skull model.

    \item A total number of $q=\noHeads$ heads (\noHeadCTs~extracted skin
        surfaces from CT scans and \noHeadScans~heads from 3D surface scans)
        were used to generate a head model.

    \item A total number of $r=\noCTs$ corresponding skulls and skin surfaces
        extracted from CT scans were used to build the FSTT statistic.

\end{enumerate}

%-----------------------------------------------------------------------------

\subsection*{Generating a parametric skull model}

In order to generate a parametric skull model we need to establish the
relationship between the different skulls from our database. For this purpose,
we register a single template skull to each skull individually. This template
model has to be a volumetric tetrahedral mesh in order to accurately represent
the solid nature of a bony skull. We therefore converted a surface triangle
mesh of a skull\footnote{Based on
\url{www.turbosquid.com/3d-models/3d-human-skull/691781}} to a volumetric
tetrahedral mesh. Our template skull model, shown in \Fig{Fig1}, consists of
$m\approx70\,\mathrm{k}$ vertices, whose positions we denote by $\set{S} =
\lbrace \vec{s}_1,\dots, \vec{s}_m \rbrace$. Tetrahedra $T(\set{S})$ are built
by connecting four vertices each, and the set of all tetrahedra is denoted as
$\set{T}=\set{T}(\set{S})$. The vertices $\set{S}$ and tetrahedra $\set{T}$
constitute the tetrahedral mesh of our skull template.

The \emph{fitting process} comprises the following two main stages for an input
skull with vertex positions $\set{P} = \lbrace \vec{p}_1,\dots, \vec{p}_M
\rbrace$:
\begin{enumerate}
    \item A global rigid transformation that coarsely aligns the input skull to
        the template skull. The registration starts with the fast global
        registration approach presented in~\cite{Zhou2016}, followed by a
        refinement step using the well known Iterative Closest Point (ICP)
        algorithm~\cite{Besl1992}.

    \item A fine registration of the template skull to the input skull, which
        consists of several non-rigid transformation steps, computed by
        minimizing the energy (inspired by~\cite{Deuss2015})
        \begin{equation}
        E(\set{S}) = \Efit\of{\set{S}} +
        \lreg\Ereg\of{\set{S}_\text{prev},\set{S}}
        \label{eqn:skull_total_energy} \end{equation} consisting of a fitting
        term $\Efit$ and a regularization term $\Ereg$.

\end{enumerate}
In the non-rigid step, the \emph{fitting term} $$
\Efit\of{\set{S}} = \frac{1}{\sum_{c \in \set{C}} w_c} \sum_{c \in \set{C}} w_c
\norm{ \vec{s}_c - \vec{f}_c }^2 \
$$ penalizes the squared distance between a vertex on the template skull
$\vec{s}_c$ and its corresponding point $\vec{f}_c$, which is a point on or
close to the mesh of the input skull. The factor $w_c \in [0,1]$ is a
per-correspondence weight, which controls the influence of the various
correspondences, such as points on the inner or outer skull surface.

The \emph{regularization term}
$$
\Ereg\of{\set{S}_\text{prev},\set{S}} = \sum_{T \in \set{T}} \left(\mathrm{vol}(T(\set{S}))- \mathrm{vol}(T(\set{S}_\text{prev}))\right)^2
$$
penalizes geometric distortion of the template skull during the fitting.
$\set{S}_\text{prev}$ represents the vertex positions of the previous
deformation state, while $\set{S}$ stands for the current (to-be-optimized)
positions. The function $\mathrm{vol}(T)$ denotes the volume of tetrahedron
$T$.  Thus, the regularization term penalizes the change of volume of tetrahedra. The non-rigid deformation starts with rather stiff material settings and successively softens the material during the registration process (by reducing $\lreg$).

During the various non-rigid transformation steps we use different strategies
to define the correspondences $\set{C}$. First, correspondences are determined
by the \emph{hierarchical ICP} approach described in~\cite{Gietzen2017}, and we
register hierarchically subdivided parts of the template skull to the input
skull using individual similarity transformations. This results in several
small pieces (e.g., the eye orbit) that are well aligned to the input skull.
Based on the correspondences found in this step the whole template skull is
registered towards the input skull.  In subsequent deformation steps, we
estimate the correspondences in a closest vertex-to-vertex manner, where we
only consider vertices lying in high curvature regions, additionally pruning
unreliable correspondences based on distance and normal
deviation~\cite{Gietzen2017}. In the final non-rigid transformation steps, when
the meshes are already in good alignment, we use vertex-to-surface-point
correspondences. These correspondences are determined considering all vertices employing a two-step search: First,
we search for vertex-to-vertex correspondences from the input skull to the
template skull, pruning unreliable correspondences based on distance and normal
deviation.  Second, we search for correspondences from the computed
corresponding vertices on the template towards the input skull. This second
step is computed in vertex-to-surface-point manner, this time pruning only
large deviation between the vertex and surface normal.

The described two-way correspondence search prevents tangential distortions of
the fitted template skull and can handle artifacts in the input skulls, e.g.,
artifacts in the teeth region due to metallic restorations. Additionally, it
makes our registration process robust against the porous bony structure caused
by low resolution of the CT scan or the age of the subject. To further prevent
mesh distortions we additionally use a release step, where the undeformed
template is deformed towards the current deformed state using only preselected
points of interest (for further details see~\cite{Gietzen2017}).

In order to analyze the accuracy of our skull registration process, we
evaluated the fitting error by computing the distance for all vertices of the
facial area (which covers all predefined landmarks) of an input skull towards
the fitted template model. The mean fitting error for all \noSkulls~fitted
skulls is below \SI{0.5}{mm}.

\begin{figure}[htbp]
	\centering
	\includegraphics[width=\textwidth]{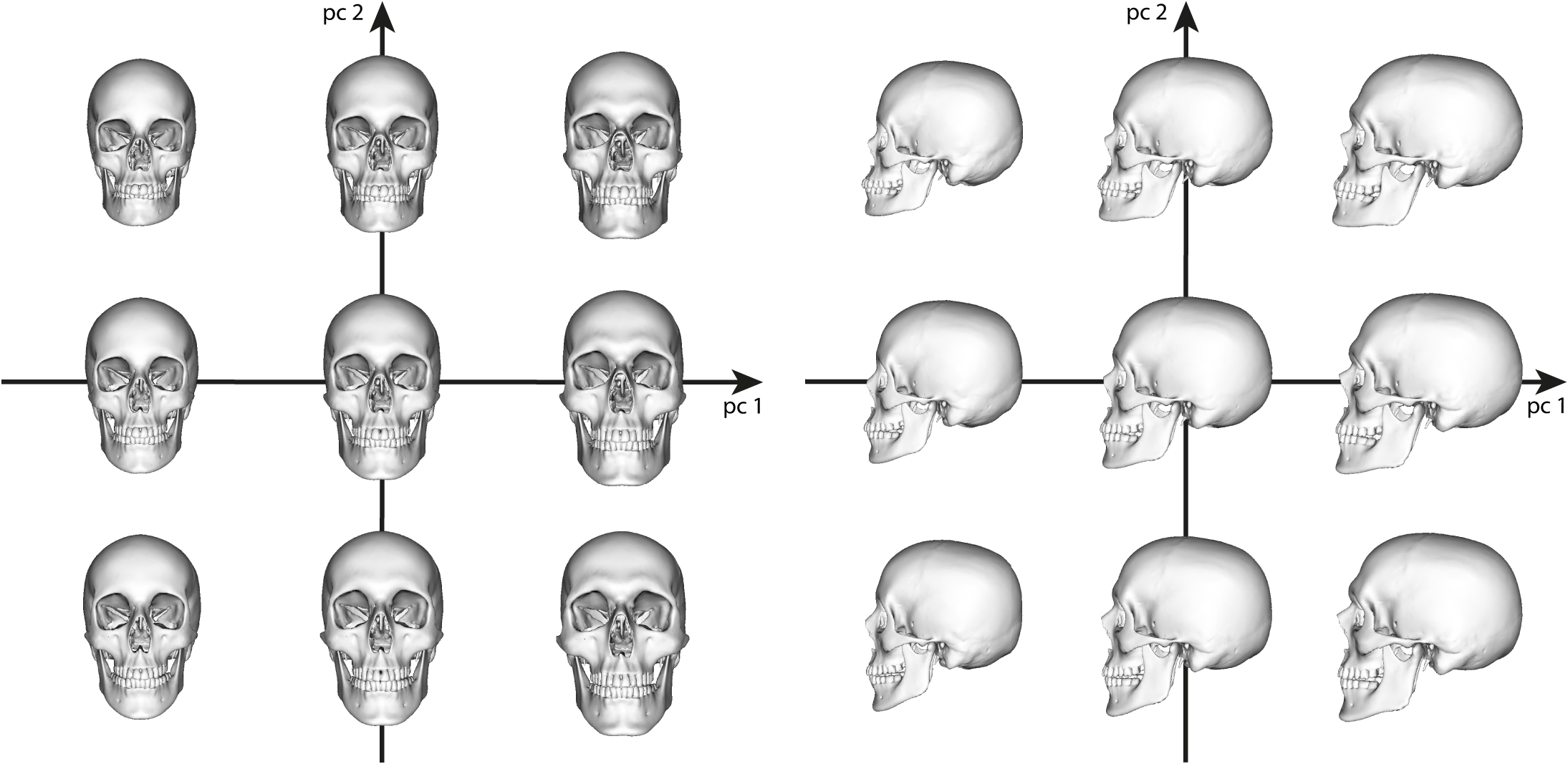}
	\caption{{\bf Skull variants along the two principal components with the largest eigenvalues.}
	We visualize $\mean{s} + \alpha_1 \vec{u}_1 + \alpha_2 \vec{u}_2$, where $\alpha_{i}=a_i \cdot \sigma_i$, $i=1,2$, is the weight containing the standard deviation $\sigma_i$ to the corresponding eigenvector $\vec{u}_i$, and the factor $a_i \in \{-2,0,2\}$.}
	\label{fig:Fig2}
\end{figure}

Stacking the vertex coordinates of each fitted skull into column vectors
$\vec{s} = (x_1,y_1,z_1,$ $\ldots,x_m,y_m,z_m)^\top$ we can apply PCA to the set of fitted skulls (after
mean-centering them by subtracting their mean $\mean{s}$).  This results in a
matrix $\mat{U} = [\vec{u}_1, \ldots, \vec{u}_{p-1}]$ containing the principal
components $\vec{u}_i$ in its columns. A particular skull $S$ in the PCA space
spanned by $\mat{U}$ can be represented as
\begin{equation}
    S(\vec{a}) = \mean{s} + \mat{U} \vec{a} ,
    \label{eqn:skullPCAModel}
\end{equation}
where $\vec{a} = \left( \alpha_1, \ldots, \alpha_{p-1} \right)^\top$ contains
the individual weights of the principal components of $\mat{U}$. The parametric
skull model \eqref{eqn:skullPCAModel} can be used to generate plausible skull
variants as a linear combination of the principal components, which is
depicted exemplarily for the first two main principal components in
\Fig{Fig2}.

We finally select 10 landmarks on the parametric skull model that are used to guide the head fitting process in the automatic forensic facial reconstruction (see detailed explanation in the section on head fitting).

%-----------------------------------------------------------------------------

\subsection*{Generating a statistic of facial soft tissue thickness}

\begin{figure}[htbp]
	\centering
	\includegraphics[width=\textwidth]{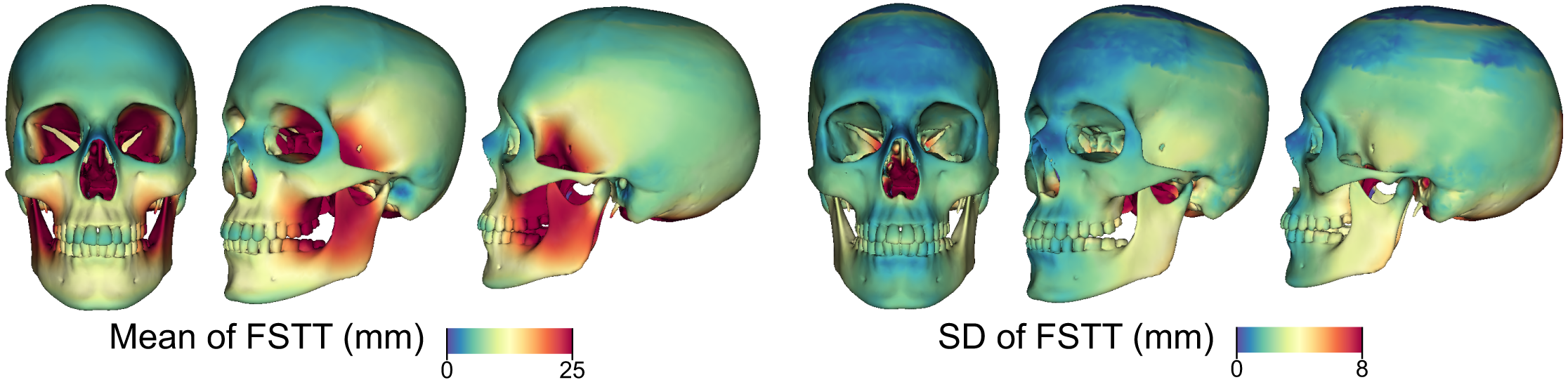}
	\caption{{\bf Statistic of the FSTT on a mean skull.} Mean and standard deviation of FSTT computed from the \noCTs\ CT scans.}
	\label{fig:Fig3}
\end{figure}

In a \emph{statistical evaluation process} the distances between \noCTs\
corresponding skulls and heads extracted from the CT scans are measured.  To
this end, we determine for each vertex of a fitted skull the shortest distance
to the surface of the extracted skin surface~\cite{Aspert2002}. Finally, the
mean and standard deviation of the FSTT are computed per vertex.  \Fig{Fig3}
shows the mean skull $\mean{s}$ with color-coded mean and standard deviation of
the obtained FSTT.

\begin{figure}[htbp]
	\centering
	\includegraphics[width=\textwidth]{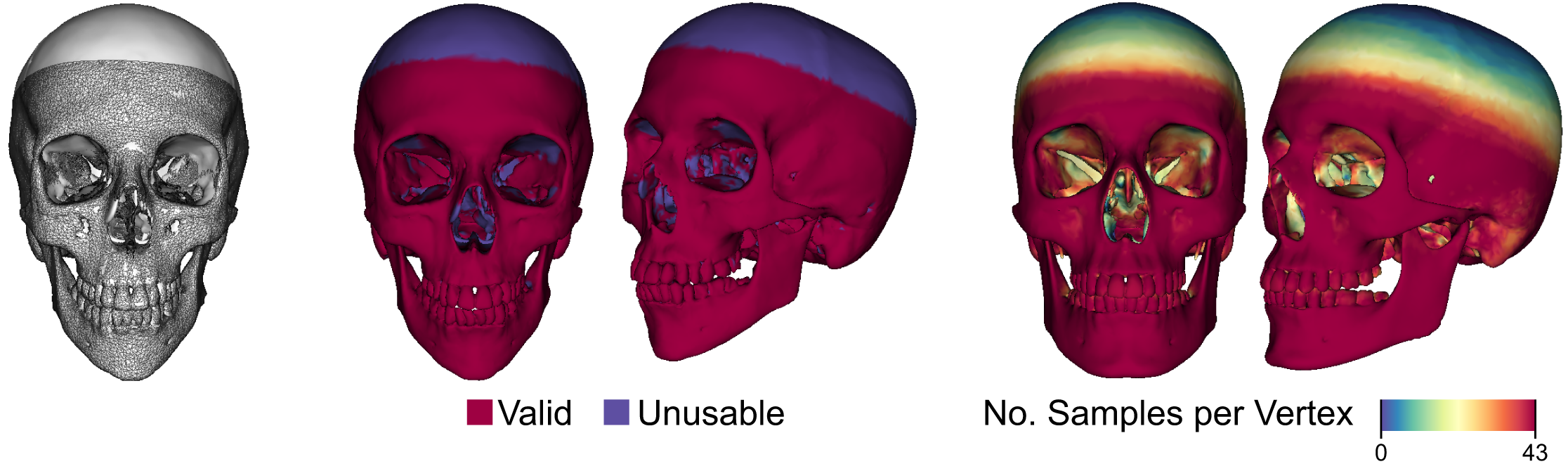}
	\caption{{\bf Basis for the statistical evaluation of the FSTT.}
	From left to right: Example of a fitted skull (white) and corresponding extracted skull (black wireframe), validation mask (corresponding to left), number of samples used for all vertices in the statistic of FSTT in \Fig{Fig3}.}
	\label{fig:Fig4}
\end{figure}

To obtain the FSTT data we often register our complete template skull to
\emph{partial} input skulls, which, for instance, have holes in the bony
structure or a missing upper part of the calvaria. \Fig{Fig4} (left) shows an
example of our template skull fitted to a partial skull extracted from CT data.
To avoid bias caused by false FSTT measurements, we validate if a vertex of a
fitted skull corresponds to a surface point on the corresponding extracted
partial skull. We exclude all vertices of the former whose distance to the
latter is larger than a given threshold (\SI{2}{mm} in our implementation). This
results in the validation mask depicted in \Fig{Fig4} (center), which
is used for the statistical evaluation. The number of FSTT measurements used
for a particular vertex in our statistic is visualized in \Fig{Fig4}
(right).  The facial skull is covered predominantly by all \noCTs~samples,
whereas the upper part of the calvaria is covered by a few samples only.

\begin{figure}[hp]
    \centering
    \includegraphics[width=\textwidth]{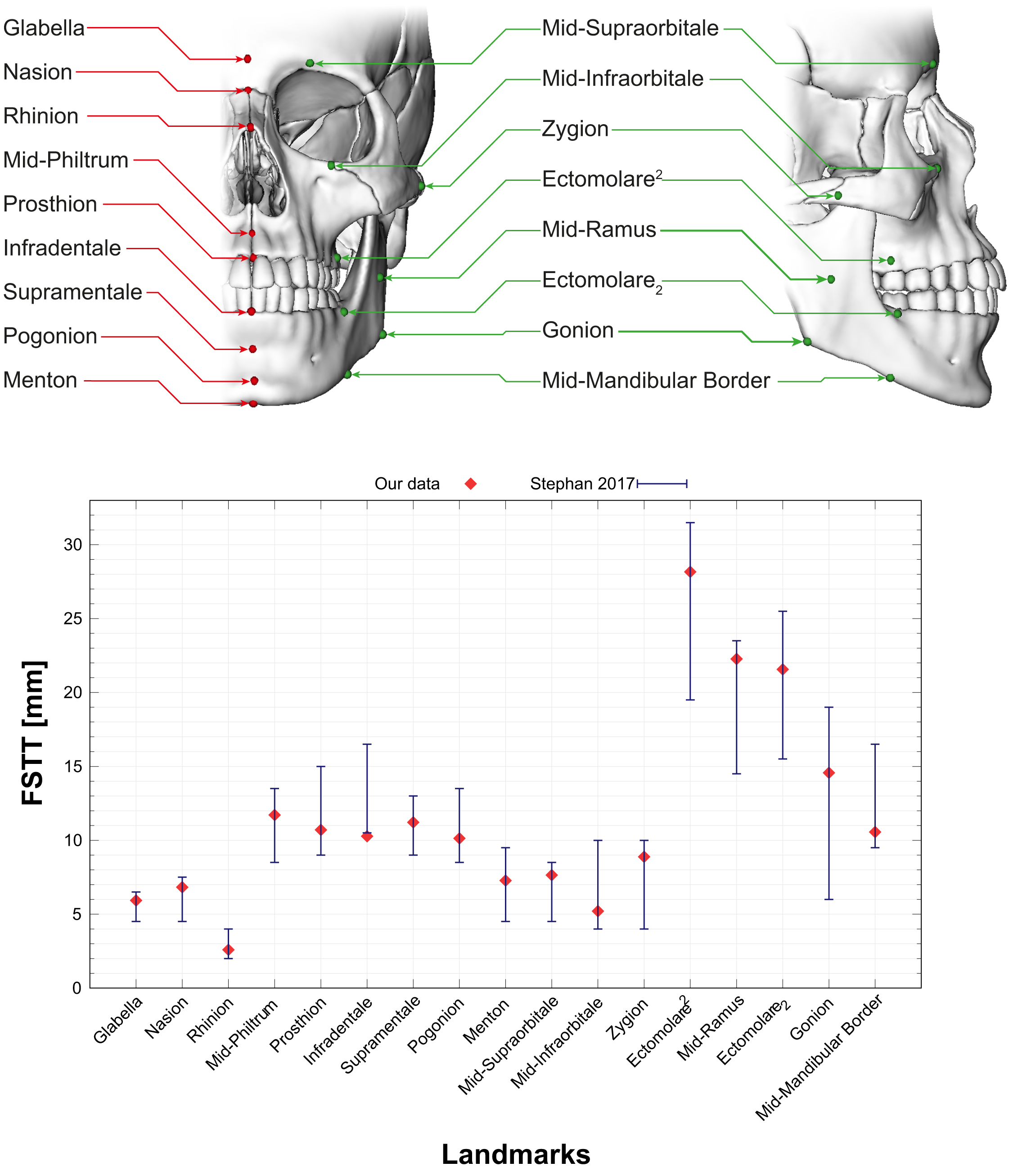}
    \caption{{\bf FSTT for commonly used midline and bilateral landmarks.}\\
    Landmarks defined by~\cite{Caple2016} as produced by our method (red dots)
in relation to pooled data from a recent meta-analysis~\cite{Stephan2017}
(weighted mean $\pm$ weighted standard deviation as blue error bars).}
    \label{fig:Fig5}
\end{figure}

The generated FSTT statistic is based on \noCTs~different subjects (26 males
and 17 females) with a mean age of 28 years.  \Fig{Fig5} presents
the computed FSTT (see \Fig{Fig3}) at some landmarks commonly
used in forensic reconstruction~\cite{Caple2016}. Our results for these
landmarks fit well into the range presented in~\cite{Stephan2017}.

%-----------------------------------------------------------------------------

\subsection*{Generating a parametric head model}

Similar to the skull model, we generate the parametric head model by fitting a template head to head scans of real subjects, which establishes correspondence between them, and then perform statistical analysis using PCA. For model generation we employ the skin surfaces extracted from the \noHeadCTs\ CT scans used for building the FSTT statistics (26 male, 17 female). However, since for some CT scans the nose tip or the upper part of the calvaria are cropped, we bootstrap the model generation by first fitting the template head to a set of \noHeadScans\ optical surface scans (20 male, 19 female) that represent complete heads. We generate a preliminary PCA model from these complete surface scans and use it to fit to the incomplete CT scans, where it fills the missing regions in a realistic manner. The final PCA model is then built from the template fits to all \noHeads\ scans.

In the following, a head scan (extracted from CT or generated through optical scan) is represented by its point set $\set{Q} = \lbrace \vec{q}_1, \dots, \vec{q}_N \rbrace$.  Since the head models are skin surfaces only, our head template is a surface triangle mesh consisting of $n \approx 6\,\mathrm{k}$ vertices with positions $\set{H} = \lbrace \vec{h}_1, \dots, \vec{h}_n \rbrace$, as shown in \Fig{Fig1}. The template fitting process consists of two stages, similar to the skull fitting:
\begin{enumerate}

\item We first optimize scaling, rotation, and translation of the template model to align it to the point set $\set{Q}$ by minimizing the sum of squared distances between points $\vec{q}_c$ on the point set $\set{Q}$ and their corresponding points $\vec{h}_c$ on the template model $\set{H}$ using ICP~\cite{Besl1992}.

\item After this coarse initialization, we perform a fine-scale non-rigid registration to update the vertex positions $\set{H}$, such that the template model better fits the points $\set{Q}$. Following the approach of~\cite{Achenbach2017}, we minimize a non-linear objective function
\begin{equation}
    \label{eqn:head_total_energy}
    E\of{\set{H}} = \Efit\of{\set{H}} + \lreg \Ereg\of{\set{H}_\text{prev}, \set{H}}.
\end{equation}

\end{enumerate}

% fitting energy
The \emph{fitting term} $\Efit$ penalizes squared distances between points $\vec{q}_c$ on the point set $\set{Q}$ and corresponding points $\vec{h}_c$ on the template model $\set{H}$:
\begin{equation}
    \label{eqn:head_fitting_energy}
    \Efit\of{\set{H}} =
    \frac{1}{\sum_{c \in \set{C}} w_c}
    \sum_{c \in \set{C}} w_c \norm{ \vec{h}_c - \vec{q}_c }^2 .
\end{equation}
The set of correspondences $\set{C}$ consists mostly of \emph{closest point correspondences}, which we construct by finding for each scan point $\vec{q}_c \in \set{Q}$ its closest surface point $\vec{h}_c$ on the template model, and which we filter by pruning unreliable correspondences based on distance and normal deviation thresholds. To allow for more precise fits, we extend these closest point correspondences by 70 \emph{facial landmarks} in the face region, on the ears, and on the lower jaw. These landmarks are manually selected on the template model and on all scans to be fitted (note that this manual work is necessary during model generation only). The per-correspondence weights $w_c$ are used to give the landmarks a higher weight than the closest point correspondences, and to assign a lower weight to surface regions that are not supposed to be fitted closely (e.g., hairs for surface scans or CT artifacts due to teeth restorations).

% deformation energy
The \emph{regularization term} $\Ereg$ penalizes the geometric distortion of the undeformed model $\set{H}_\text{prev}$ (the result of the previous rigid/similarity transformation) to the deformed state $\set{H}$.  Since the head template is a surface mesh, we employ a discrete surface deformation model that  minimizes bending, discretized by the squared deviation of the per-edge Laplacians
\begin{equation}
\label{eqn:regularization_energy}
\Ereg\of{\set{H}_\text{prev}, \set{H}} =
\frac{1}{\sum_{e \in \set{E}} A_e}
\sum_{e \in \set{E}}
A_e \norm{\laplace^e\vec{h}(e) - \mat{R}_e \laplace^e\vec{h}_\text{prev}(e)}^2.
\end{equation}
Here, $A_e$ is the area associated to edge $e$, and $\mat{R}_e$ are per-edge rotations to best-fit deformed and undeformed Laplacians (see \cite{Achenbach2015} for details). In the spirit of non-rigid ICP~\cite{Achenbach2017} we alternatingly compute correspondences and minimize \eqref{eqn:head_total_energy}, starting with a rather stiff surface that is subsequently softened (by reducing $\lreg$) to allow for more and more accurate fits.  Whenever $\lreg$ is decreased, we also update the rest state $\set{H}_\text{prev}$ by the current deformed state $\set{H}$.

% preliminary PCA model
From the \noHeadScans\ fits to the complete optical surface scans we construct a preliminary parametric head model. Similar to the skull model generation, we stack the vertex positions of each fitted head $\vec{h}=\left(x_1,y_1,z_1,\ldots,x_n,y_n,z_n\right)^\top$ and compute a PCA model of dimension $d$ ($d=30$ in our case), such that we can write
\begin{equation}
    \label{eqn:headPCAModel}
    H(\vec{b}) = \mean{h} + \mat{V} \vec{b},
\end{equation}
where $\mean{h}$ is the mean head, $\mat{V}$ is the matrix containing the principal components in its $d$ columns, and $\vec{b} = (\beta_1,\ldots,\beta_{d})$ contains the PCA parameters representing the head.

% fit PCA model to incomplete CT scans
With the preliminary PCA model at hand, we can now fit the head template to the incomplete skin surfaces extracted from CT scans, where regions of missing data are filled realistically by the PCA model. Fitting to a point set $\set{Q}$ amounts to additionally optimizing the PCA parameters $\vec{b}$ during the initial rigid/similarity transformation step. To this end, we minimize squared distances of corresponding points, with a Tikhonov regularization ensuring plausible weights:
\begin{equation}
\label{eqn:headPCAFit}
    \Epca(\vec{b}) \;=\;
    \frac{1}{\sum_{c \in \set{C}} w_c}
    \sum_{c \in \set{C}} w_c
    \norm{ \mean{h}_c + \mat{V}_c \vec{b} - \vec{q}_c }^2 +
    \frac{\ltik}{d}
    \sum_{k=1}^{d} \left( \frac{\beta_k}{\sigma_k} \right)^2 .
\end{equation}
In the fitting term, $\mat{V}_c$ and $\mean{h}_c$ are the rows of $\mat{V}$ and $\mean{h}$ representing the point $\vec{h}_c$ corresponding to $\vec{q}_c$, that is $\vec{h}_c = \mean{h}_c + \mat{V}_c\vec{b}$. We use $\ltik = 1\cdot10^{-4}$ for the regularization term, where $\sigma_k^2$ is the variance of the $k$th principal component. The optimal weights $\vec{b}$ are found by solving the linear least-squares problem \eqref{eqn:headPCAFit}. In step (1) of the head fitting process we optimize for alignment (scaling, rotation, translation) and for shape (PCA weights) in an alternating manner until convergence. Step (2), the non-rigid registration, is then performed the same way as without the PCA model.

\begin{figure}[htbp]
	\centering
    \includegraphics[width=\textwidth]{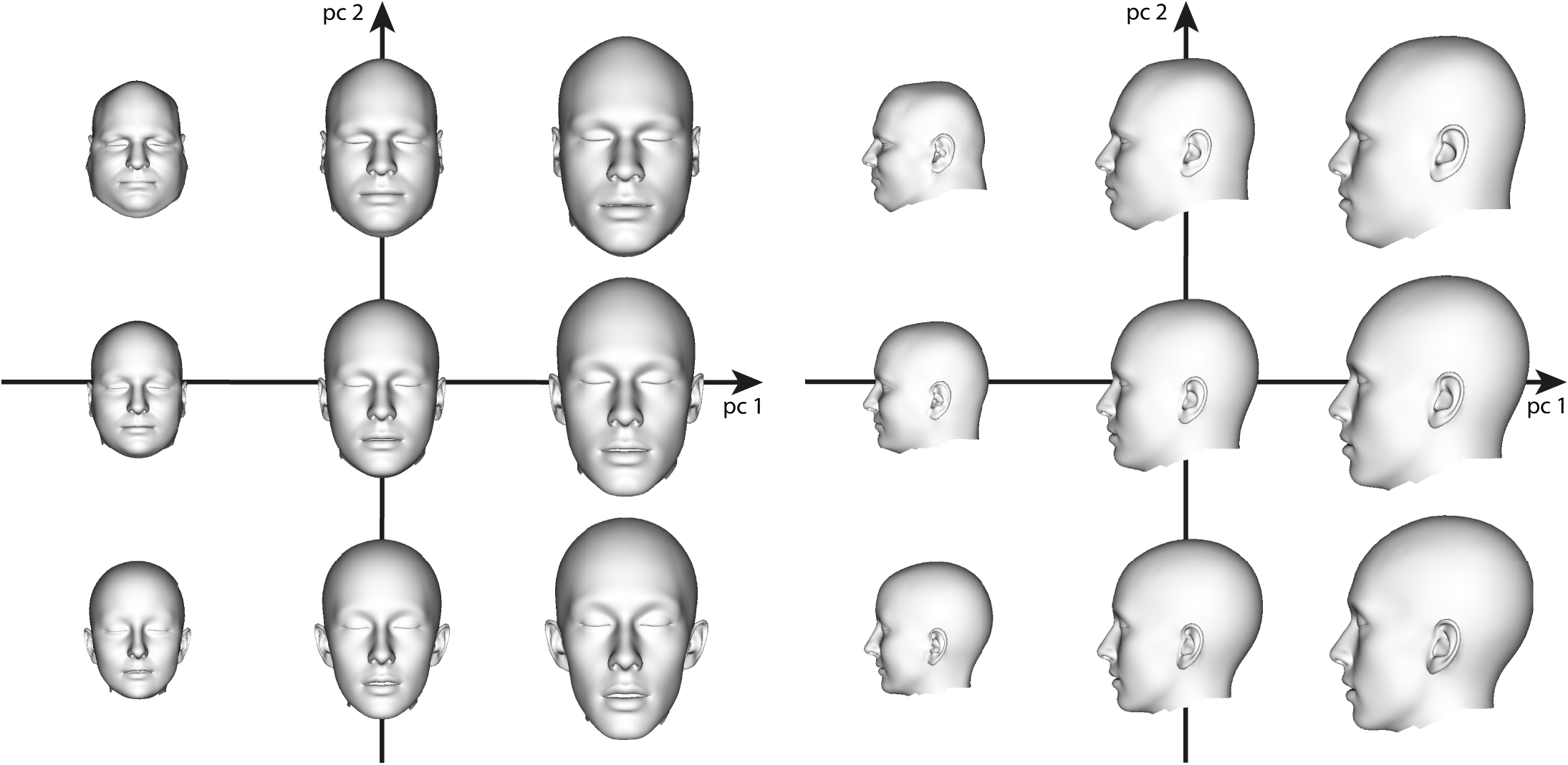}
    \caption{{\bf Head variants along the two principal components
    with the largest eigenvalues.} We visualize $\mean{h} + \beta_1 \vec{v}_1 + \beta_2 \vec{v}_2$, where $\beta_{i}=b_i \cdot \sigma_i$, $i=1,2$, is the weight containing the standard deviation $\sigma_i$ to the corresponding eigenvector $\vec{v}_i$, and the factor $b_i \in \{-2,0,2\}$.}
    \label{fig:Fig6}
\end{figure}

We finally combine the fits to the \noCTs\ CT scans and to the \noHeadScans\ surface scans into a single parametric PCA head model.  The variation of this model along the first two principal directions is shown in \Fig{Fig6}.  While the first principal component basically characterizes head size, the second principal component describes strong variation of head shape within our training data.

In order to analyze the accuracy of our head fitting process, we evaluate the RMS error for all \noHeads\ head scans:
\[
    \mathrm{rms}(\set{H}, \set{Q}) =
    \sqrt{
        \frac{1}{\sum_{c \in \set{C}} w_c}
        \sum_{c \in \set{C}} w_c
        \norm{ \vec{h}_c - \vec{q}_c }^2
    }.
\]
This is similar to \eqref{eqn:head_fitting_energy} and measures the distance between corresponding point pairs from $\set{H}$ and $\set{Q}$. Depending on our input data, we weight down regions that should not be fitted closely (hairs, CT artifacts), such that these regions do not influence the error measure too much.  Averaging this error over all \noHeads\ scans gives an overall fitting error of \SI{0.19}{mm}.  Note that we prune unreliable correspondences above a distance threshold of \SI{2}{mm}, which therefore are not considered for error evaluation.  However, since the overall fitting error is an order of magnitude smaller, it is not significantly influenced by this pruning.

As done before for the parametric skull model, we also manually select 10
corresponding landmarks on the parametric head model, which are used for the
automatic forensic facial reconstruction.

%%%%%%%%%%%%%%%%%%%%%%%%%%%%%%%%%%%%%%%%%%%%%%%%%%%%%%%%%%%%%%%%%%%%%%%%%%%%%%

\section*{Automatic forensic facial reconstruction}

\begin{figure}[htbp]
    \centering
    \includegraphics[width=\textwidth]{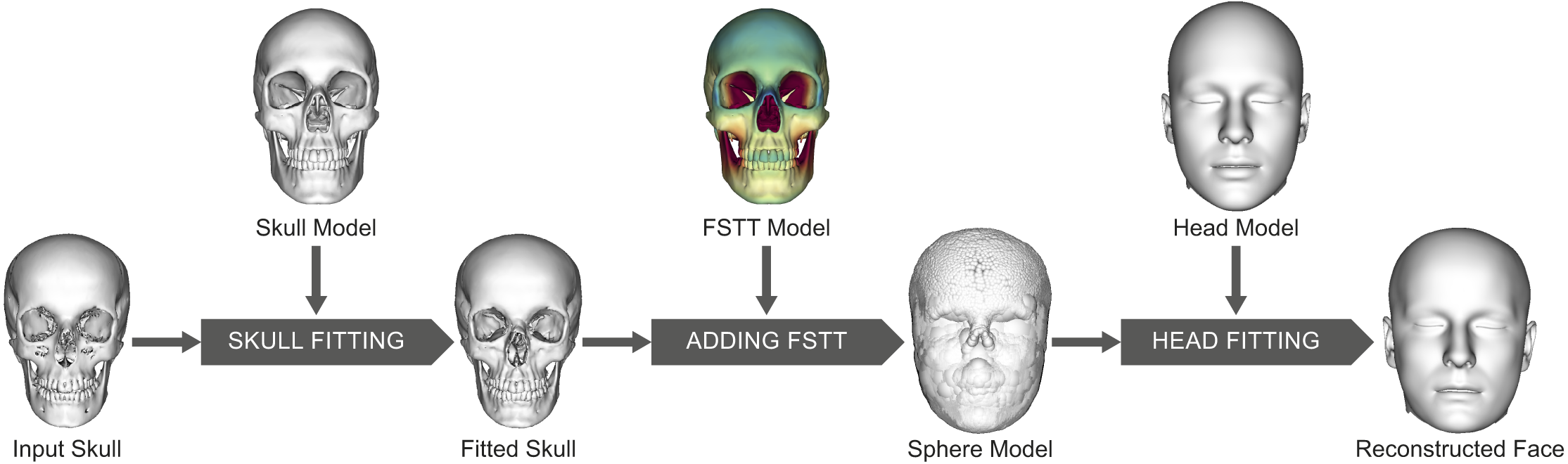}
    \caption{{\bf Processing steps of the automatic forensic facial reconstruction.} The reconstruction of a face from a given input skull utilizing the generated parametric skull model, the statistic of FSTT, and the parametric head model.}
    \label{fig:Fig7}
\end{figure}

Our automatic forensic facial reconstruction process is based on the generated parametric skull model, the statistic of FSTT, and the parametric head model, described in the previous sections. In the following, we use an anonymized CT scan of a female subject with an age of 21 years to demonstrate the quality of our forensic facial reconstruction.  This CT scan was not used for constructing the parametric skull model, head model, or FSTT statistic.  The reconstruction process runs in three steps as shown in \Fig{Fig7} and is explained in the following sections.

%-----------------------------------------------------------------------------

\subsection*{Skull fitting}

Given scanned skull remains as input, the \emph{skull fitting} process is very similar to the registration process described in the section about generating the parametric skull model.  The main difference is that we are finally able to utilize the generated parametric skull model \eqref{eqn:skullPCAModel} as a starting point for the subsequent deformation steps.  First, we compute a shape-preserving transformation which aligns the parametric skull model to the given skull by using the global registration approach presented in~\cite{Zhou2016}.  To further optimize the alignment we search for reliable point correspondences $\set{C}$ between the given skull and the parametric skull model and compute the optimal scaling, rotation, and translation in closed form~\cite{Horn1987}.  After optimizing the alignment, we continue with optimizing the shape.  Similar to the PCA fitting of heads~\eqref{eqn:headPCAFit} we are looking for the coefficient vector $\vec{a}$ of the parametric skull model \eqref{eqn:skullPCAModel} with
\begin{equation}
	\min_{\vec{a}}
    \frac{1}{\abs{\set{C}}}
    \sum_{c \in \set{C}}
    \norm{ \mean{s}_{c} + \mat{U}_c \vec{a} - \vec{p}_c }^2 +
    \frac{\ltik}{d} \sum_{k=1}^{d} \left(\frac{\alpha_k}{\sigma_k}\right)^2,
	\label{eqn:skullPCAFit}
\end{equation}
where $\ltik = 1\cdot10^{-3}$, $\sigma_k^2$ is the variance of the $k$th principal component $k$ of the skull model and $d$ is the number of employed PCA components. Optimization for alignment and shape is alternated until convergence, and before each optimization (alignment or shape) we recompute point correspondences $\set{C}$.  After this initialization, we continue with non-rigid registration by minimizing \eqref{eqn:skull_total_energy}.

%-----------------------------------------------------------------------------

\subsection*{Adding facial soft tissue thickness}

\begin{figure}[htbp]
	\centering
	\includegraphics[width=\textwidth]{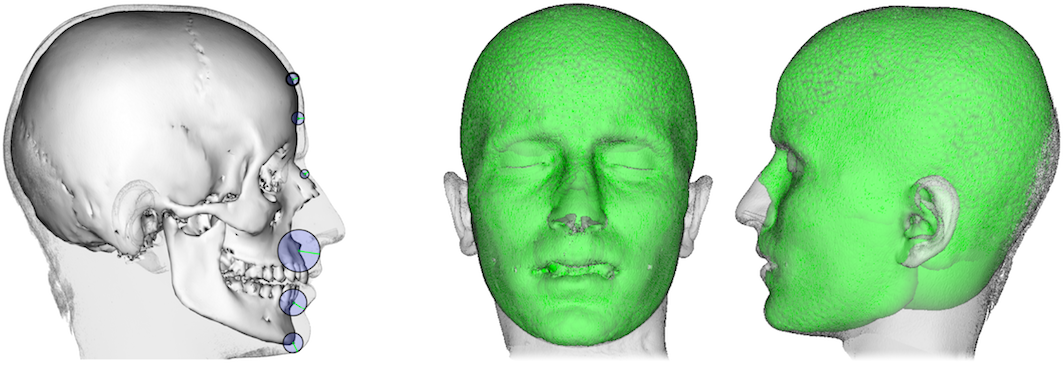}
    \caption{{\bf FSTT for a given individual visualized as sphere model.} At each skull vertex a sphere with radius of the actual FSTT value from the ground truth data set is drawn. From left to right: Some example spheres for points on the midline, union of all spheres (in green) with original skin surface as overlay.}
	\label{fig:Fig8}
\end{figure}

Next we assign FSTT values based on our FSTT statistic to the fitting result for a given skull. An important advantage of our approach is that out FSTT statistics only contains \emph{scalar} FSTT values without a particular measurement direction, such as skull normal or skin normal, since these directions are hard to determine in a robust manner due to noise or fitting errors.  
In our case the measured skin position, which is the closest point on the skin surface for a vertex of the skull, is located on a sphere centered at the skull vertex with radius being the corresponding FSTT value. 
\Fig{Fig8} (left) shows a side view of the FSTT measurement results for few preselected points on the midline.

Knowing both the skull and the skin surface for a subject allows the
computation of the \emph{actual} FSTT. \Fig{Fig8} (center and right) shows an
overlay of the extracted skin surface and the union of all spheres centered at
the skull vertices and having as radii the appropriate FSTT values, which we
call the \emph{sphere model}.  The depicted sphere model is based on the exact
FSTT of this subject and provides a visually good approximation of the real skin
surface.  Certainly, since nose and ears do not have a directly underlying bony
structure, this method does not provide this kind of information.  Approaches
for prediction of nasal morphology, such as \cite{Kahler2003, Rynn2010}, give
some hints about the nose, e.g., the approximated position of the nose tip, but
do not really create an individual nose shape for a particular subject. In a
real application scenario the age, sex and ancestry of the individual are
derived from its skeleton remains and a disaggregated FSTT statistic is used
for reconstruction. In our case the sample size is too small to build specific
FSTT statistics, so as an approximation we simply build the sphere model based
on the mean of our general FSTT statistics (cf.\ \Fig{Fig7}).

%-----------------------------------------------------------------------------

\subsection*{Head fitting}

\begin{figure}[htbp]
\centering
\includegraphics[width=0.9\textwidth]{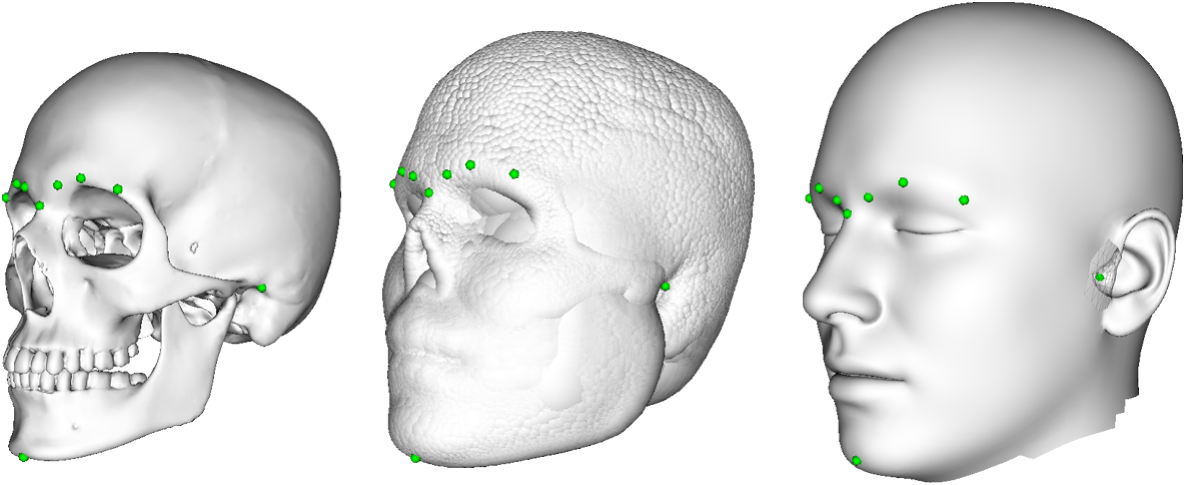}
\caption{{\bf Landmarks for the automatic facial reconstruction.} From left to right: Mean skull with preselected landmarks, sphere model based on mean FSTT with projected landmarks, and mean head with preselected landmarks. The landmarks consist of two midline landmarks and four bilateral landmarks, which are selected once on the parametric skull and head model after model generation. The landmarks are based on the proposed nomenclature of~\cite{Caple2016}: \emph{nasion} and \emph{menton} (from craniometry) and \emph{mid-supraorbitale} and \emph{porion} (from craniometry) as well as \emph{ciliare lateralis} and \emph{ciliare medialis} (from capulometric) and their corresponding counterparts on skull respectively skin surface.}
\label{fig:Fig9}
\end{figure}

Given a specific sphere model, the next step is to derive a facial profile from this data. For this purpose we deform our parametric head model to the (under-specified) sphere model. The fitting procedure is very similar to the generation of our parametric head model.  Similar as before, we initially align the sphere model with the parametric head model.  However, this time the landmarks on the fitted skull, which have been selected during the skull model generation, are projected automatically onto the surface of the sphere model as depicted in \Fig{Fig9}.

The projected landmarks give us robust correspondences on the parametric head
model. They are automatically determined and replace the manually selected
landmarks used during model generation. We start by optimizing scaling,
rotation, and translation, as well as PCA parameters based on the set of
landmarks. This initialization is followed by a fine-scale non-rigid
registration based on landmarks and closest point correspondences between the
parametric head model and the given sphere model.

While this process is very similar to the model generation phase, it differs in
the following point: We use the per-correspondence weights $w_c$ in the fitting
energy \eqref{eqn:head_fitting_energy} to give points on the outer surface of
the sphere model more influence than points in the interior, since the former
can be considered as an approximation to the skin surface that we intend to
fit. To this end, we first identify if a point $\vec{q}_c$ on the sphere model
is outside from its corresponding point $\vec{h}_c$ on the head template by
checking $\vec{n}_c^{\top} (\vec{q}_c - \vec{h}_c) \geq 0$, where $\vec{n}_c$
is the normal vector of $\vec{h}_c$.  For such correspondences, we set $w_c = 1
+ 10^8 \cdot \norm{ \vec{h}_c - \vec{q}_c } / B$, where $B$ is the bounding box
size of model.

As mentioned before, nose and ears do not have a directly underlying bony
structure. Thus the sphere models do not provide any data for such regions.
Utilizing a parametric head model allows the reconstruction of nose and ears in
a statistical sense, i.e., as an element related to the underlying PCA space.

%-----------------------------------------------------------------------------

\subsection*{Generating plausible head variants}

\begin{figure}[htbp]
\centering
\includegraphics[width=\textwidth]{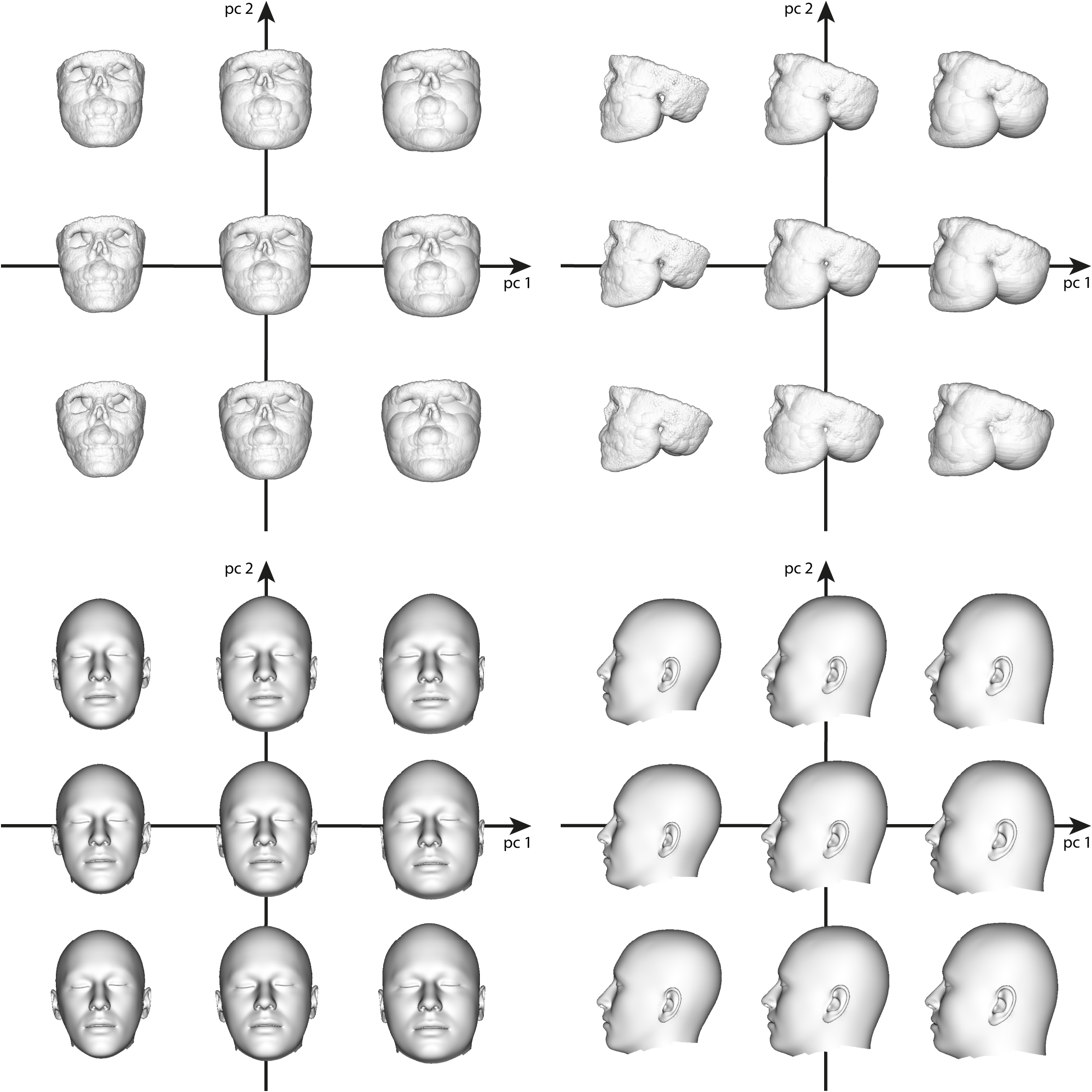}
\caption{{\bf Variants of plausible FSTT distributions for the anonymized given skull.} Top: Partial sphere model variants along the two principal components with the largest eigenvalues: We visualize $\mean{t} + \gamma_1 \vec{w}_1 + \gamma_2 \vec{w}_2$, where $\gamma_{i}=c_i \cdot \sigma_i$, $i=1,2$, is the weight containing the standard deviation $\sigma_i$ to the corresponding eigenvector $\vec{w}_i$, and the factor $c_i \in \{-2,0,2\}$. Bottom: Head model fitted to these partial sphere models.}
\label{fig:Fig10}
\end{figure}

The simplest method for facial reconstruction is to fit the template head to a sphere model based on the \emph{mean} of the FSTT statistics.  However, this approximation will rarely match a specific subject. To get a reliable FSTT diversification for an individual, we again adopt the PCA approach creating a parametric FSTT model
\begin{equation}
    \textrm{FSTT}(\vec{c}) = \mean{t} + \mat{W} \vec{c}
    \label{eqn:fstt_pca-model}
\end{equation}
where $\mean{t}$ is the mean FSTT, $\mat{W}$ contains the principal components of the FSTT, and $\vec{c} = (\gamma_1,\ldots,\gamma_{r-1})$ contains the PCA parameters. Using this parametric FSTT model, we can create plausible FSTT variants for the given input skull. Since the CT scans used for the statistic of FSTT are mostly missing the upper part of the calvaria, the FSTT values obtained in this area are mainly very large and invalid. Thus we omit this area for the construction of our parametric FSTT model \eqref{eqn:fstt_pca-model}, which results in partial sphere models. \Fig{Fig10} (top) depicts a subset of the partial sphere models along the two principal components with the largest eigenvalues for the given input skull.

Our head fitting process described above can be applied to the partial sphere
models without special adjustments. As depicted in \Fig{Fig10} (bottom) our
approach is able to generate plausible head variants based on the corresponding
sphere models in \Fig{Fig10} (top). As we are using a parametric model of the
complete head, the missing parts like nose, ears and especially the skin
surface above the calvaria, are reconstructed in a statistical sense, i.e., as
an element related to the underlying PCA space.

%%%%%%%%%%%%%%%%%%%%%%%%%%%%%%%%%%%%%%%%%%%%%%%%%%%%%%%%%%%%%%%%%%%%%%%%%%%%%%

\section*{Discussion and conclusion}

In this paper we presented an automated method based on a parametric skull
model, a parametric head model, and a statistic of FSTT for reconstructing the
face for a given skull. The models we are using were derived from head CT scans
taken from an existing CT image repository and from 3D surface scans of real
subjects. Our approach has three main outcomes: (i) a dense map of FSTT (i.e.,
a soft tissue layer), (ii) a visual presentation of a statistically probable
head based on a statistic of FSTT and a parametric head model, and (iii) a
method to generate plausible head or face variants, respectively.

The main advantage of our approach over landmark-based FSTT measurements (see
references in~\cite{Stephan2017}) is the density of the FSTT map without the
need of error-prone normal information.  For any vertex of the parametric skull
model a FSTT value can be derived from the statistic of FSTT.  It is important
to note that the statistical evaluation of the FSTT is fully automatic without
any manual interaction.  This is different from other FSTT assessments based on
CT data, which often still rely on error-prone manual measurements (see, e.g.,
\cite{Cha2013}). The fully automated method introduced here can help to
generate a more accurate database in the future, largely overcoming the
accuracy issues well-known for manual, landmark-based FSTT assessments
\cite{Stephan2008}. However, as our method is based on CT scans, it is still prone to typical artifacts and gravity effects due to supine patient position.
Although our statistic of FSTT so far is generated from
only \noCTs~CT scans, the data we derived (\Fig{Fig5}) clearly indicate good
agreement with data just recently published in a meta-analysis
\cite{Stephan2017}. If enough appropriate CT scans are available, rapid
processing by means of an automated pipeline can aid the creation of a large
statistical database. It seems most likely that methods such as the one
introduced here constitute the future for the generation of statistical models
from 3D medical imagery. Therefore, enlarging the database will be part of our
future work to generate a more precise statistic.

A statistic of FSTT plays a significant role in facial
approximation~\cite{Stephan2008} and is also an integral part of modern
orthodontic treatment planning \cite{Ackerman1999,Cha2013}. For forensic
reconstruction, it forms the basis for further steps in the reconstruction
process.  While traditional facial reconstruction methods rely on manual
clay-based sculpturing, which strongly depends on the operator's artistic
abilities and subjective interpretation \cite{Wilkinson2010}, automated methods
based on a dense statistical model can help to overcome such
ambiguities~\cite{Wilkinson2010}. The advantage of our approach in comparison
to other automated methods
\cite{Turner2005,Tu2007,Romeiro2014,Shui2016,Shui2017} is that our facial
reconstruction process is fully automated. The only manual steps done in our
approach are during the model generation processes. As mentioned before, our
statistic of FSTT is independent of the measurement direction and thus we
utilize sphere models in the reconstruction process. Therefore, error-prone
strategies such as averaging over normal vectors to define a measurement
direction are completely avoided. Moreover, our parametric FSTT model allows us
to create plausible head variants in a statistical sense, which do not require
any prior knowledge.

Subsequently, future work will concentrate on merging the two pathways
(parametric skull and head model) by integrating all statistical information
into one combined model. This model could then be used for various purposes,
such as forensic applications, demonstrations for medical procedures, yet also
for realistic animations in movies.

\begin{figure}[htbp]
\centering
\includegraphics[width=\textwidth]{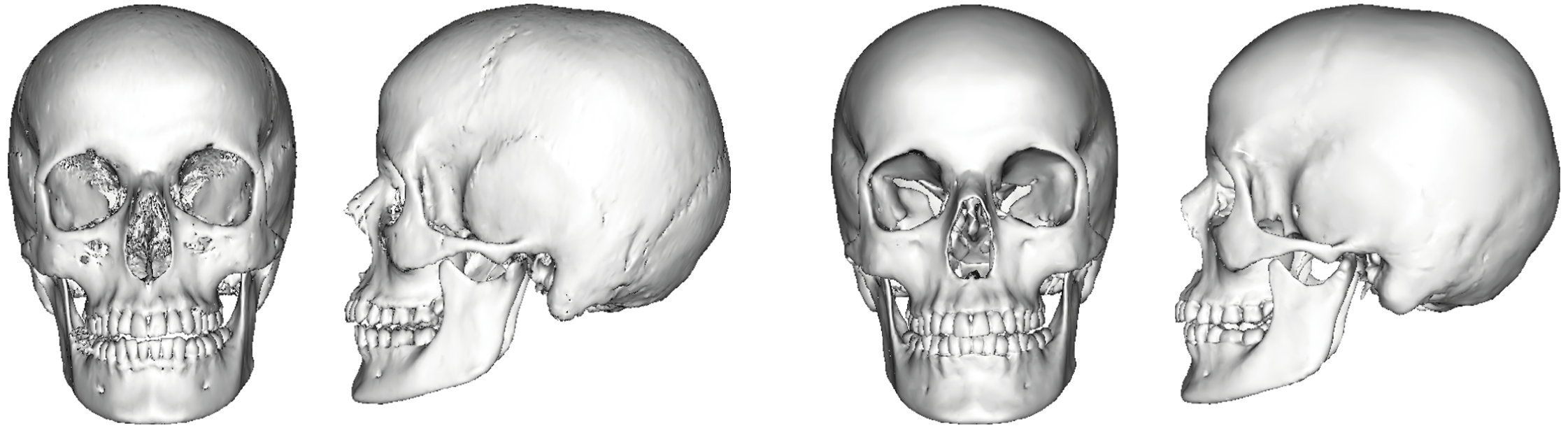}
\caption{{\bf Skull fitting results for a given skull.} Extracted skull from CT (left) and fitted skull (right).}
\label{fig:Fig11}
\end{figure}

\begin{figure}[htbp]
\centering
\includegraphics[width=\textwidth]{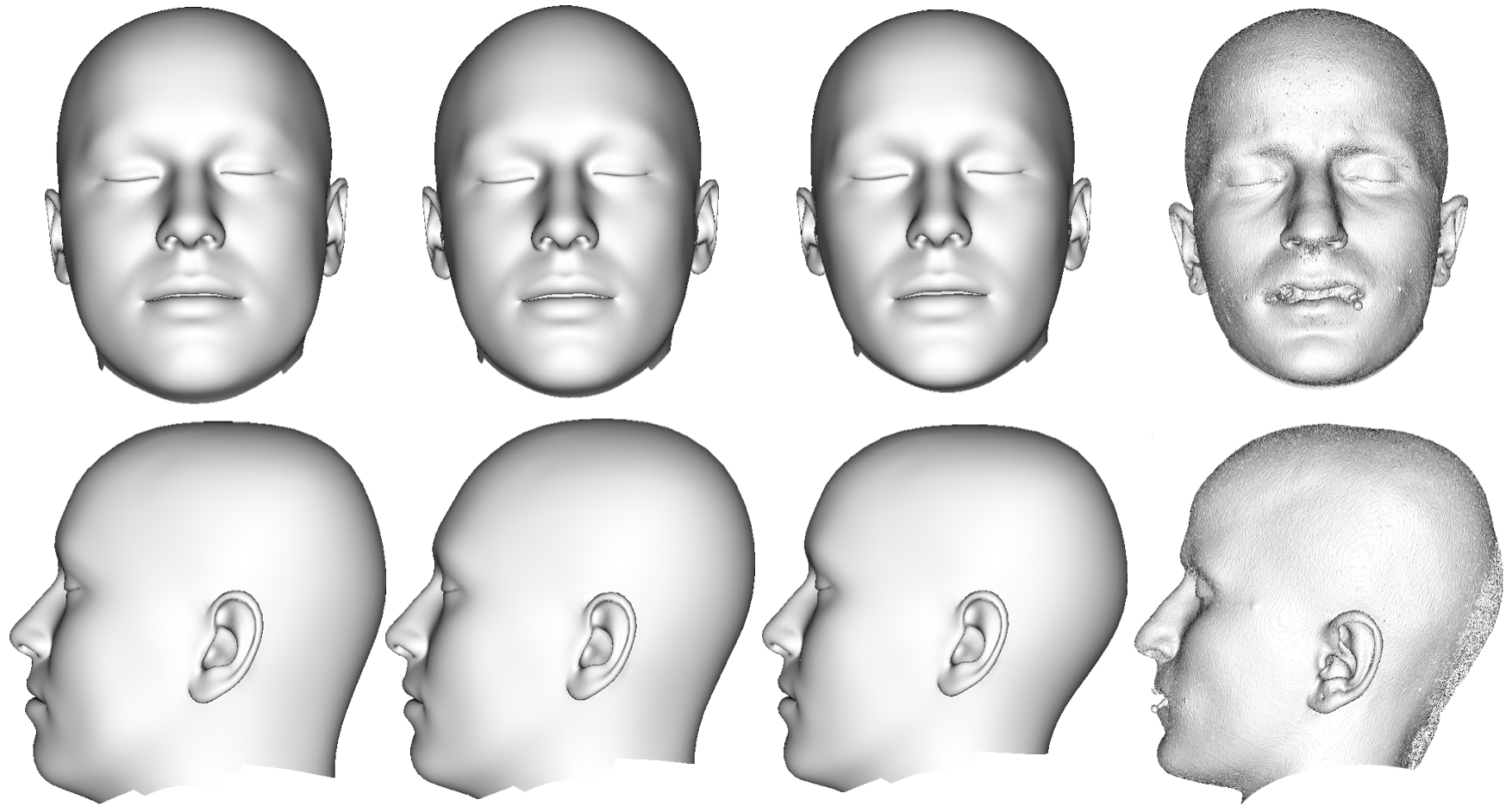}
\caption{{\bf Comparison of head fittings and extracted skin surface from CT.} From left to right: Fitted head to sphere model based on mean FSTT, fitted head to sphere model based on $\mean{t} - 1.5\sigma_1\vec{w}_1$, fitted head to sphere model based on original FSTT, extracted skin surface from CT.}
\label{fig:Fig12}
\end{figure}

In conclusion, the automated technique suggested in this paper aids recognition
of unknown skull remains (e.g.\ see \Fig{Fig11}) by providing statistical
estimates derived from a CT head database and 3D surface scans. By creating a
range of plausible heads in the sense of statistical estimates, a ``visual
guess'' of likely heads can be used for recognition of the individual
represented by the unknown skull.  Compared to clay-based sculpturing, which
depends on the ability of the operator, our method provides a good
approximation of the facial skin surface in a statistical sense (see
\Fig{Fig12}).  Nevertheless, the quality of the reconstruction depends on the
sample size of the statistic.  In order to use additional descriptive factors
(e.g., age, sex, ancestry, weight), a larger sample size representing the variance of
each of the factors is required. We thus aim to enlarge our skull and head
database to further elaborate on the methods introduced here.

%%%%%%%%%%%%%%%%%%%%%%%%%%%%%%%%%%%%%%%%%%%%%%%%%%%%%%%%%%%%%%%%%%%%%%%%%%%%%%

\clearpage
\section*{Acknowledgments}

The results from this study are part of the research projects ``Kephalos'' and ``KogniHome'' funded by the Federal Ministry of Education and Research (BMBF).  The work was also supported by the Cluster of Excellence Cognitive Interaction Technology ``CITEC'' (EXC~277) at Bielefeld University, which is funded by the German Research Foundation (DFG).  We gratefully acknowledge the Department of Diagnostic and Interventional Radiology, University Medical Center of the Johannes Gutenberg University Mainz, Germany for providing us with the DICOM data of the human heads.

%%%%%%%%%%%%%%%%%%%%%%%%%%%%%%%%%%%%%%%%%%%%%%%%%%%%%%%%%%%%%%%%%%%%%%%%%%%%%%

\bibliography{literature}

%%%%%%%%%%%%%%%%%%%%%%%%%%%%%%%%%%%%%%%%%%%%%%%%%%%%%%%%%%%%%%%%%%%%%%%%%%%%%%
\end{document}